\documentclass[runningheads]{llncs}

\usepackage[T1]{fontenc}
\usepackage{graphicx}
\begin{document}
\title{Call2Instruct: Automated Pipeline for Generating Q\&A Datasets from Call Center Recordings for LLM Fine-Tuning}
\titlerunning{Call2Instruct}
%
\author{Alex Echeverria, Sávio Salvarino Teles de Oliveira, Fernando Marques Federson}


%
\authorrunning{ Alex Echeverria et al.}
%
\institute{ Instituto de Informática -- Universidade Federal de Goiás\\
Goiânia -- GO -- Brazil\\
\email{federson@ufg.br}}
\maketitle              
\begin{abstract}
The adaptation of Large-Scale Language Models (LLMs) to specific domains depends on high-quality fine-tuning datasets, particularly in instructional format (e.g., Question-Answer - Q\&A). However, generating these datasets, particularly from unstructured sources such as call center audio recordings, poses a significant challenge due to the noisy and disorganized nature of the data. This paper presents a solution to this challenge by offering an end-to-end automated pipeline for generating Q\&A instructional datasets from such recordings. The methodology developed comprises sequential steps of audio processing (including diarization, noise removal and automatic transcription), textual processing (cleaning, normalization, and anonymization), semantic extraction of customer demands and attendant responses using vector embeddings, and matching via semantic search to form the final Q\&A pairs. As a result, the complete pipeline was successfully implemented, generating a dataset specifically formatted for Instruct Fine Tuning. The practical value and feasibility of the generated dataset were substantiated and functionally demonstrated through the successful fine-tuning of an LLM model (based on Llama 2 7B). The conclusion of the paper states that the proposed approach is viable for converting unstructured conversational data from call centers into valuable resources for training LLMs. This development has the potential to open up avenues for creating more effective AI systems for Q\&A tasks in the customer service domain. The developed codes have been made publicly available to promote reproducibility and future research.

\keywords{Large Language Models \and Q\&A Dataset Generation \and Call Centers.}
\end{abstract}
\section{Introduction}
Large Language Models (LLMs) have established themselves as a technology with a significant impact on the field of artificial intelligence, generating great interest in both academic and corporate environments~\cite{zhao23}. Among the areas of research that involve LLMs are the methods to train them and, fundamentally, the techniques to adapt and specialize them in specific tasks or domains, a process known as fine-tuning~\cite{zhao23}. An essential component of successful fine-tuning is the availability of high-quality instructional datasets, specific to the desired task, often in Question-Answer (Q\&A) format~\cite{liu24,schimanski24}.

However, creating these datasets represents a considerable challenge, especially when seeking to take advantage of abundant but inherently unstructured data sources. Audio recordings of call center interactions, for example, contain valuable information about real customer demands and the corresponding solutions, but require complex processing to be transformed into a useful format. Despite substantial advances in both general LLM fine-tuning techniques~\cite{hu21,ouyang22} and in isolated stages of call center data processing - such as automatic speech recognition (ASR) and the application of Natural Language Processing (NLP) for textual analysis - a critical gap remains: the development of integrated, automated systems capable of efficiently transforming these raw sources into instructional datasets, ready for supervised LLM training.

In fact, recent studies demonstrate the effectiveness of LLM models applied to specific tasks such as summarizing customer conversations~\cite{Nivethitha24} and extracting insights from transcripts via document retrieval~\cite{crisp24}. However, these approaches remain fragmented, addressing steps such as audio capture, transcription, linguistic processing, or data adaptation in isolation, without offering an end-to-end solution for generating instructional datasets. Therefore, the ability to automatically transform audio recordings into structured Question-Answer pairs, ready for the fine-tuning of LLMs in Q\&A tasks, remains an area with room for contributions. 

This paper proposes and details an automated pipeline designed specifically for generating instructional datasets, in Question-Answer format, from call center audio recordings. The approach presented encompasses a complete flow of operations: it starts with obtaining and preprocessing the audio (including channel separation, noise removal, and transcription), moves on to textual preprocessing (cleaning and anonymization of transcribed data), uses semantic search and embeddings to extract customer demands and relevant responses from agents, and culminates in the construction of a properly formatted dataset for Instruct Fine-Tuning LLMs in the Q\&A task. The system developed therefore aims to offer a viable, automated solution to the challenge of creating specific, high-quality datasets from raw conversational data. In addition, unlocking the potential of these conversational data would make it possible to improve the efficiency and quality of customer service and capture valuable operational knowledge that is dispersed and underutilized.

The structure of this article is as follows. Section 2 provides a concise overview of the work related to the proposed pipeline. Section 3 describes the developed approach in detail. In Section 4, we present the experimental scenarios, the data set, and an analysis of the results for each stage. Finally, Section 5 concludes the article and outlines potential directions for future research.

\section{Related Work}
The development of our proposed pipeline, Call2Instruct, relies on processing unstructured call center recordings, making Automatic Speech Recognition (ASR) a critical initial stage. However, accurately transcribing audio in this domain poses significant challenges, as extensively documented in the literature. Call center environments are often characterized by background noise and channel distortions inherent to telephone or VoIP communications, which severely degrade ASR performance ~\cite{parra21,fernandez22,suziki16}. Moreover, their conversational nature frequently leads to issues such as speaker overlap and multiple participants, further complicating transcription and speaker attribution ~\cite{suziki16,chetupalli21,do20}. These combined factors typically result in high Word Error Rates (WER), negatively impacting the utility of transcripts for downstream tasks ~\cite{mamou06,obaidah24}.

To address these difficulties, research has proposed various strategies to improve ASR robustness in such demanding conditions. These include the development of specialized acoustic models, often combined with denoising techniques ~\cite{parra21}, as well as approaches to handling overlapped speech ~\cite{suziki16}. Integration of ASR with speaker diarization or segmentation methods has also shown promise for managing multi-speaker scenarios and correctly attributing utterances to customers or agents ~\cite{chetupalli21,do20}. Additional studies explore mitigating transmission artifacts, such as packet loss, through data augmentation techniques ~\cite{fernandez22}. Given the inherent difficulty, some works even focus on performing information retrieval directly on high-WER transcripts ~\cite{mamou06}, while others contribute benchmarks to evaluate progress in this challenging domain ~\cite{obaidah24}. Together, these efforts underscore the complexity of producing reliable textual data from call recordings, a prerequisite for high-quality Q\&A dataset generation.

Beyond transcription, raw textual output introduces further obstacles for downstream processing. Transcripts from spontaneous speech often lack standard textual structure, including punctuation and capitalization, limiting the readability and performance of the subsequent components of NLP ~\cite{behre22}. Additionally, such transcripts frequently contain disfluencies such as hesitations, repetitions, and self-corrections, which obscure semantic content ~\cite{lin20}. To address this, researchers have developed automatic text normalization techniques. Sequence-to-sequence models and fine-tuned large pre-trained models such as BERT have shown promise for punctuation restoration and truecasing ~\cite{fu22,melamud21}. Some approaches jointly tackle these tasks with disfluency detection to produce cleaner and more structured transcriptions from the raw ASR output ~\cite{lin20,cui23}.


A major challenge in working with call center data is ensuring privacy and compliance with regulations such as GDPR (General Data Protection Regulation in the European Union) and Brazil’s LGPD. Transcripts often contain Personally Identifiable Information (PII), which requires robust anonymization before use in analysis or model training. Identifying and redacting PII is particularly challenging given the informal and unstructured nature of conversations and the potential transcription errors introduced by ASR systems ~\cite{kaplan20,kurata12}. Named Entity Recognition (NER) serves as a foundational technique to automatically detect PII in text ~\cite{kaplan20,hassan18}. Significant research has been devoted to improving the robustness of NER for noisy telephone transcripts through specialized training, model distillation, and architectures designed to handle ASR ambiguity ~\cite{kaplan20,fu22,kurata12}. Recent work also explores using LLMs for data augmentation to improve de-identification performance ~\cite{dhingra24}, as well as developing dedicated tools to streamline anonymization pipelines ~\cite{mazzarino23}.

With preprocessed and anonymized conversational text, the focus shifts to extracting relevant information to form meaningful Q\&A pairs. Understanding the structure and semantics of the dialog is crucial. Techniques such as Dialogue Act (DA) classification — which labels each utterance’s communicative function ((e.g., question, answer, request) — and Intent Recognition — which identifies the main goal or need of the user within the conversation — are widely studied for this purpose ~\cite{gao22,moradizeyveh22,xu22}. These tasks often rely on advanced models such as graph neural networks or transformers ~\cite{xu22,moradizeyveh22}, and are essential for identifying customer queries and corresponding agent responses, particularly in complex multi-turn dialogues typical of call center scenarios ~\cite{webb05}.


However, simply identifying question and answer segments may be insufficient. Responses may be delayed or phrased differently, requiring a semantic link between customer demand and agent resolution. Semantic search techniques based on text embeddings address this issue ~\cite{trivedi18}. Models like BERT and its variants generate dense vector representations that capture meaning beyond surface form, enabling similarity matching ~\cite{guo19}. These techniques have proven effective in related domains such as retrieving similar questions in community Q\&A forums ~\cite{guo19,othman19} or matching user tickets with known solutions in enterprise systems ~\cite{trivedi18,xie19}.

Building on these components, our work contributes to the broader domain of automatic Q\&A dataset generation. Although various methods have been proposed to generate Q\&A pairs from structured documents ~\cite{pham24}, knowledge graphs ~\cite{khapra17}, or clean text corpora such as Wikipedia ~\cite{lee20,akyon21}, these approaches often rely on relatively clean and well-structured input. In contrast, generating instruction-style Q\&A datasets from noisy, spontaneous conversational recordings - like those of call centers - involves a unique combination of challenges: robust ASR, effective normalization, rigorous anonymization, accurate dialogue understanding, and reliable semantic matching. While components have been addressed individually, the development of end-to-end automated pipelines like Call2Instruct, designed to transform raw call recordings into valuable fine-tuning resources for LLMs in the customer service domain, remains an area with significant room for contribution, which this paper aims to address.

\section{Proposed Method}
\begin{figure}
\includegraphics[width=\textwidth]{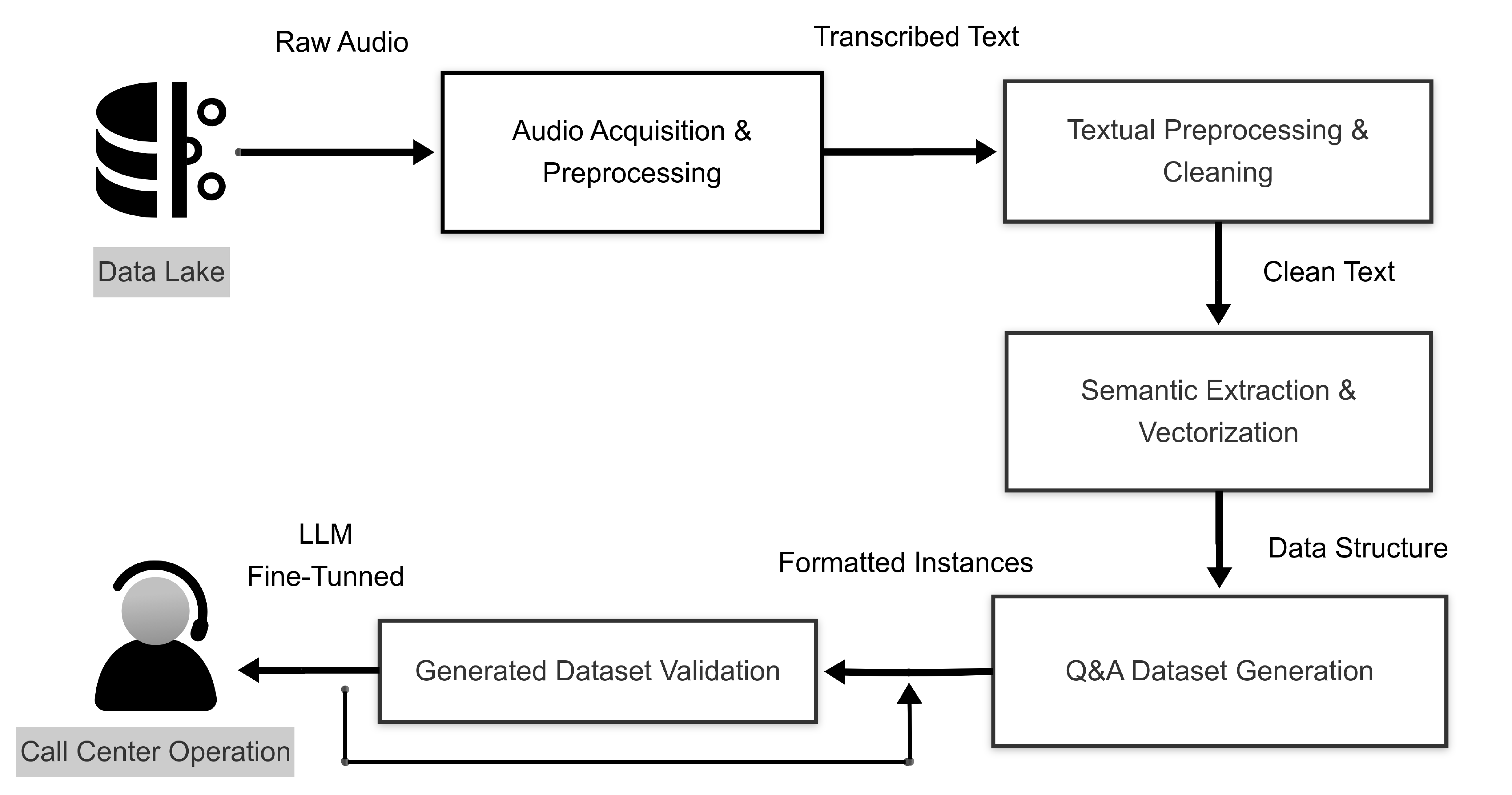}
\caption{Conceptual overview of the Call2Instruct automated pipeline.} \label{fig1}
\end{figure}

The Call2Instruct pipeline is designed to automatically transform raw call center conversations into high-quality instruction-style question-answer (Q\&A) datasets. The process ensures data privacy, semantic coherence, and training readiness for natural language understanding models. The methodological process, conceptually illustrated in Figure~\ref{fig1}, was structured into five sequential and interdependent data transformation stages, progressing from raw audio to the final Question-Answer (Q\&A) dataset format, culminating in a validation of the utility of the generated dataset. The stages are described in the following.

\subsection{Audio Acquisition and Preprocessing}



The pipeline begins with the acquisition of recorded customer service calls, typically stored in compressed audio formats in a call center interaction repository (data lake). This stage involves converting the audio to a uniform format and sampling rate, followed by denoising and segmentation into speaker turns when needed. These procedures ensure that the recordings are suitable for automatic speech recognition (ASR) and downstream text processing. The process includes the collection of audio recordings, the application of channel separation techniques (diarization) to isolate the agent's and customer's speech into distinct audio channels for subsequent speaker labeling, and the utilization of denoising algorithms and filtering to mitigate background noise and identify/remove nonrelevant segments such as Interactive Voice Response (IVR) audio. Following these preprocessing steps, Automatic Speech Recognition (ASR) is performed to convert the preprocessed audio to text. The result of this stage was a set of raw textual transcripts, with speech associated with its respective speaker (customer or agent).

\subsection{Textual Preprocessing and Cleaning}




Once the audio is transcribed using a robust ASR engine, the resulting raw text is cleaned and normalized. This includes punctuation restoration, correction of transcription artifacts, and sentence segmentation. Additionally, sensitive information is identified and anonymized using placeholder tokens (e.g., \texttt{<NAME>}, \texttt{<ACCOUNT\_ID>}), preserving privacy without compromising contextual relevance. Therefore, this stage focuses on improving the text's quality and suitability for downstream processing by identifying and correcting common errors originating from the ASR process, removing disfluencies (hesitations, repetitions) and other artifacts of spontaneous speech to enhance text clarity, and implementing mechanisms to detect and remove or mask Personally Identifiable Information (PII) present in transcripts, ensuring privacy and compliance with data protection regulations. The result is a clean and anonymized textual corpus.

\subsection{Semantic Information Extraction and Vectorization}



This phase aims to identify and structure the core information from conversations (customer demands and agent responses) semantically that can support the generation of instructional Q\&A examples. This includes detecting user intents, identifying relevant dialogue acts, and extracting utterance-response pairs. To support this, sentence embedding models are applied to represent utterances in a semantic vector space, enabling similarity-based filtering and pair matching for high-quality dialogue snippets. The executed steps involve the development of methods to automatically detect text segments corresponding to the main customer requests or questions (demands) and the answers or solutions provided by the agents (responses). The identified segments were then rewritten to ensure greater objectivity, clarity, and removal of excessive markers of orality, making the primary intent of each demand and response explicit. Finally, vector representations (embeddings) for the rewritten demands and responses were generated using pre-trained language models (e.g., OpenAI's text-embedding-ada-002 model) and stored in a vector database (e.g., Elasticsearch) to enable efficient semantic similarity-based searches.

\subsection{Question-Answer (Q\&A) Dataset Generation}



Using the paired utterances and contextual dialogue, instruction-style data points are instantiated in the input and answer format. Each pair is enriched with natural prompts (e.g., 'What did the customer ask?' or 'What was the agent's recommendation?') generated using prompt templates. This stage aligns with instruction-tuning practices in modern LLM training, aiming to teach models how to respond to user instructions based on real conversational data. Therefore, using the vector database, the final creation of the dataset proceeded by performing a semantic similarity search for each customer demand (represented as a vector) within the vector database to retrieve a set of the most semantically relevant agent responses. The customer demands were then paired with the retrieved responses, forming instances of Question (customer demand) and Answer (agent response), and these Q\&A pairs were structured in a specific format suitable for Instruct Fine-Tuning of LLMs, including natural language instructions describing the Q\&A task.

\subsection{Generated Dataset Validation}




To ensure that the generated data meets quality and privacy standards, a validation stage is implemented. A fine-tuning experiment is conducted following an implicit evaluation until the defined metrics are met. This includes automatic checks for coherence, redundancy, and completeness, as well as the verification that no sensitive information was leaked post-generation. A sample of the dataset is also manually reviewed to validate instruction diversity, response relevance, and fluency. Only validated entries are retained for inclusion in the final dataset. To evaluate the efficacy and quality of the automatically generated dataset, a fine-tuning experiment was conducted, using the instruction-formatted dataset to perform Instruct Fine-Tuning on a pre-trained LLM (e.g. Llama 2 7B~\cite{touvron23}). Although the successful execution of fine-tuning serves as functional validation in this study, a formal evaluation methodology would typically include metrics assessing the post-fine-tuning model's performance on relevant Q\&A tasks (e.g., BLEU, ROUGE, METEOR, or human evaluation) to quantitatively or qualitatively demonstrate the success of the process.



%
%
\section{Experiments and Results}
The implementation of the proposed methodology (detailed in Section 3) culminated in the development of an automated pipeline for generating instructional Question-Answer (Q\&A) datasets from a call center audio recordings in particular. This section details the results obtained in each of the crucial stages of the Call2Instruct pipeline.

\subsection{Processed Data and Initial Transcripts}

The pipeline's starting point was the acquisition and preprocessing of raw audio recordings from a call center interaction repository (stored as buckets on Amazon Web Services - AWS). As outlined in Section 3.1 (Audio Acquisition and Preprocessing), this initial phase comprised the following sub-steps and their respective outcomes:

\begin{itemize}
    \item Data Acquisition and Preparation: The records were initially retrieved from the AWS buckets and prepared for subsequent processing.

    \item Channel Separation (Diarization) and Noise Removal: In this specific case, the audio channels were already separated, distinguishing agent speech from customer speech. Therefore, diarization was not necessary. Subsequently, the \texttt{denoiser} framework~\cite{defossez20} was used to mitigate background noise, aiming to improve audio quality for transcription.

    \item Unwanted Segment Filtering (IVR/URA Removal): Segments corresponding to Interactive Voice Response (IVR) or automated attendant messages (Unidade de Resposta Audível - URA) were identified and removed. Given that these segments consist of prerecorded messages and menus that do not contribute to the Q\&A pairs relevant to human interaction, an empirical, unsupervised approach was developed and applied. This method involved: (i) loading the agent's audio channel (which typically contains the IVR/URA); (ii) segmenting the audio into fixed-time windows; (iii) extracting acoustic features from each window; (iv) applying K-Means clustering to group these feature windows, thereby distinguishing IVR/URA patterns from actual conversation; and (v) removing segments preceding the identified transition point from IVR/URA to human interaction.

    \item Automatic Speech Recognition (ASR): The preprocessed audio, now cleaned of initial noise and IVR/URA segments, was then fed into an ASR system. For this task, tools such as Faster Whisper and Insanely Fast Whisper were utilized. These tools leverage OpenAI's Whisper Large model, which is optimized with techniques such as low rank adaptation (LoRA), quantization, and Flash Attention~\cite{faster23,insanely23,radford23,hu22,dao22} for improved efficiency and speed in transcription.
\end{itemize}

The direct outcome of this first macro-stage was a corpus of raw textual transcripts. In these transcripts, the speech was segmented and preliminarily attributed to the respective speaker (customer or agent). 



\subsection{Cleaned and Anonymized Textual Corpus}

Following initial transcription, the raw textual data, while representing the desired conversational content, often contained various errors and noise inherent to the ASR process, which are detrimental to LLM training~\cite{zhao23,chen24}. Exploratory analysis of the initial transcripts revealed common issues such as incorrect transcriptions of speech, unexpected hallucinations (particularly in silent segments, manifested as repetitive or meaningless terms), and temporal misalignments between speaker turns.

Therefore, a textual preprocessing and cleaning stage (as outlined in Section 3.2) was implemented. This stage aimed to refine the transcripts by addressing ASR artifacts and normalizing the text. Inspired by the challenges and operators discussed in work such as Data Juicer~\cite{chen24}, which highlights the heterogeneity of data sources and the need for robust preprocessing for LLMs, specific cleaning routines were developed. While Data-Juicer offers over 50 operators, a subset of relevant principles (such as filtering repetitive phrases and normalizing numerical representations) was adapted and implemented to address the specific issues identified in our call center transcripts, such as non-standardized number transcriptions (e.g., "two zero zero" vs. "two hundred") and misspellings of specific terms or names.

Concurrently, and of paramount importance, an anonymization process was applied to identify and remove or mask Personally Identifiable Information (PII) commonly found in call center interactions. This step was critical to ensure data privacy and compliance with data protection regulations. The outcome of this phase was a cleaned and anonymized textual corpus, significantly more suitable for the subsequent semantic analysis and Q\&A pair generation. This refined dataset mitigated the impact of transcription errors and protected sensitive information, laying a more robust foundation for the next stages of the pipeline.

\subsection{Vector Database of Demands and Responses}

Following the cleaning and anonymization of the textual corpus, the subsequent stage focused on semantic information extraction and vectorization, as detailed in Section 3.3. 

The primary goal was to identify and structure the core information content of the conversations, specifically customer demands and agent responses, to support the generation of instructional Q\&A examples. To achieve this, a multistep process was employed. First, customer utterances from the cleaned transcripts were processed to explicitly state their demands or the reason for their call. This involved submitting these utterances to ChatGPT~\cite{chatgpt23} tasked with rewriting the customer's speech into a more objective and direct question format, removing typical oral markers such as prolixity or unnecessary repetitions. This rewriting step, guided by a carefully refined prompt structure that incorporates few-shot examples, proved crucial to standardizing the representation of customer needs. Initial prompt designs required adjustments, as evidenced by a validation set of 40 randomly selected transcripts where an improved prompt successfully rewrote all demands, compared to an earlier version that failed on 19 instances. A more detailed description of this prompt engineering process is available in the project's GitHub repository~\cite{GITHUB}.

Once the customer demands were reformulated and, similarly, the corresponding agent responses were identified from the transcripts, both were converted into dense vector representations (embeddings). For this task, OpenAI's text-embedding-ada-002 model~\cite{neelakantan22} was selected due to its robustness and ease of use via API. This model generates a 1536-dimensional floating-point vector for each textual input. Both the rewritten customer demands and the preprocessed agent responses underwent this embedding extraction process.

To efficiently store and query these embeddings, Elasticsearch~\cite{elasticsearch18} was used as a vector database. Each entry in Elasticsearch contained the text segment (either customer demand or agent response), the speaker's persona (customer or agent), the original audio file identifier, and its corresponding embedding vector. This setup facilitated semantic search capabilities, allowing, for instance, a customer's demand embedding to be used as a query to find the most semantically similar agent responses across the entire dataset, not limited to the original interaction. This approach, inspired by information retrieval techniques~\cite{abbasiantaeb21}, enables the discovery of relevant information between different interactions and agents.

The outcome of this stage was a structured knowledge base in which key conversational intentions were represented as vectors, prepared for semantic search and the subsequent formation of Q\&A pairs.

\subsection{Instructional Q\&A Dataset Generation}

With vectorized demands and responses stored in Elasticsearch, the pipeline proceeded to the Question-Answer (Q\&A) dataset generation phase, as described in Section 3.4. This stage leveraged the semantic search capabilities of the vector database to form meaningful instructional pairs.

For each rewritten customer demand (vectorized), a semantic similarity search was performed against the entire collection of vectorized agent responses. The objective was to retrieve a set of the most semantically relevant agent responses, regardless of their original interaction context. Initial explorations revealed that directly using the top raw retrieved responses could still include undesirable or imprecise information. Therefore, a refinement process was implemented. For each customer demand, the best N (in this implementation, N=3) semantically similar agent responses were selected as candidates. These candidate responses were then individually submitted to ChatGPT, with a prompt designed to rewrite each response to be more clear, objective, and focused on the information pertinent to the customer's demand, while discarding irrelevant details. This step aimed to improve the quality and conciseness of potential answers.

Furthermore, an additional check was performed in parallel using ChatGPT to identify any customer demands that might be "invalid" (e.g., a transcription fragment that does not constitute a genuine query). Of a population of 3120 rewritten customer demands, only a negligible number (2 instances) were flagged as potentially invalid by this process. These were noted but not immediately excluded, acknowledging the potential need for further curation in larger datasets.

The core of this stage was to synthesize a single, comprehensive answer from the multiple refined candidate responses for each customer demand. Recalling that the ultimate goal is to create Q\&A instances for LLM fine-tuning, the N refined candidate responses were again processed by ChatGPT. The LLM was instructed to read these N candidates and generate a single, coherent answer that incorporates all relevant information from the candidates, maintains clarity, and removes any residual noise or unnecessary details. This step effectively consolidated the knowledge spread across multiple relevant agent interactions into a singular, high-quality answer paired with the customer's demand.

Finally, these demand-answer pairs were structured into a specific format suitable for instruct fine-tuning of LLMs. This involved creating instances where the "question" was the rewritten customer demand and the "answer" was the synthesized agent response, often framed with natural language instructions (e.g., "Based on the customer's query, what was the recommended solution?"). The direct result of this phase was the final instructional Q\&A dataset, ready for use in fine-tuning LLMs for customer service Q\&A tasks.

\subsection{Functional Validation via Fine-Tuning}
To demonstrate the practical utility and assess the quality of the generated instructional Q\&A dataset, a fine-tuning experiment was conducted as the final validation step of the pipeline, aligned with the procedures described in Section 3.5. This involved using the dataset created by Call2Instruct to perform instruct fine-tuning on a pre-trained Large Language Model (LLM).

The dataset, derived from more than 3000 customer call recordings related to the telecommunications sector, was formatted according to principles that facilitate instruct fine-tuning, drawing insights from works such as Wei et al.~\cite{wei21} on how LLMs can be improved as zero-shot learners through instruction. This involves structuring data with natural language instructions that describe the task, enabling the model to learn the nuances of human-like command following.

For this experiment, the Llama 2 7B model~\cite{touvron23} was chosen due to its strong baseline performance and the availability of a version already fine-tuned for chat-like interactions, which bear resemblance to Q\&A tasks. The fine-tuning process itself was carried out using the Lamini platform~\cite{lamini24}. Lamini was selected for its ability to abstract away many of the low-level complexities of LLM training, allowing for a streamlined fine-tuning process (e.g., requiring only a few lines of code to initiate training on their cloud-based, auto-managed infrastructure). Although this approach offers less granular control compared to manual fine-tuning by an LLM expert, it was deemed suitable for the scope of this work, which primarily focuses on dataset preparation.

The successful execution of the fine-tuning process using the Call2Instruct-generated dataset serves as a key functional validation. The LLM was fine-tuned on practical examples of queries and responses from the telecommunications call center, resulting in a model adapted to this specific domain. Qualitative validation, performed by posing various simulated customer demands to the fine-tuned model, indicated that it had learned the call center context and could generally understand and respond to customer queries without deviating from the domain. For example, the model was able to address common customer issues based on patterns learned from the dataset. However, this qualitative assessment also revealed areas for improvement. In some instances, the fine-tuned model provided responses that were too generic or lacked the specificity required to fully resolve a nuanced query. This suggests that further enhancements could be made to both the model training regimen and, critically, the quality and diversity of the generated dataset itself, particularly in ensuring highly specific and actionable responses are well represented.

The adaptation of the Llama 2 7B model confirms that the Call2Instruct pipeline is viable to convert unstructured call center recordings into a valuable resource for training LLMs, effectively bridging the gap between raw conversational data and domain-specific AI assistants. The code developed for this fine-tuning process, along with practical examples, has been made available on the project's GitHub repository~\cite{GITHUB} to promote reproducibility.

\section{Conclusion and Future Work}

This paper has presented the development and implementation of an automated pipeline for generating instructional Question-Answer (Q\&A) datasets from audio recordings of call center interactions. The primary outcome, the successful generation of a dataset and its functional validation through the fine-tuning of the LLM Llama 2 7B, demonstrates the viability of the proposed approach for transforming raw, unstructured conversational data into valuable resources for training advanced language models. The significance of this research lies in addressing the critical bottleneck in creating high-quality, domain- or task-specific datasets by offering a systematic path to unlock the potential of inherently noisy and unstructured call center data for Instruct Fine-Tuning. The ability to automatically generate relevant Q\&A pairs from real conversations can lead to more effective and contextually aware LLMs for customer service applications. The end-to-end approach, integrating audio processing, text processing (including an important rewriting step for normalization), and semantic techniques, represents a comprehensive solution, with the public availability of the code contributing to reproducibility.

However, several important limitations are acknowledged. The quality of the final dataset is intrinsically dependent on the accuracy of each pipeline component, especially the ASR system, as transcription errors can propagate. The effectiveness of textual cleaning, anonymization, and the semantic search-based pairing methodology (which may not always retrieve the optimal answer) are also crucial. Furthermore, the validation performed was primarily functional; a rigorous quantitative evaluation of the resulting LLM's performance is necessary. The generalization of the pipeline to other domains or languages was not explored, and there is room for optimization in the prompts and algorithms used.

Building on these findings and limitations, future work will focus on several key directions. A robust quantitative evaluation framework will be implemented, including domain-specific benchmarks and human assessment. Pipeline components, particularly ASR, textual cleaning, anonymization, and Q\&A pairing methods (perhaps exploring cross-encoder models or LLMs for quality assessment), will be enhanced. The impact of the rewriting step and the use of LLMs for summarization or keypoint extraction before Q\&A generation will be further explored. Generalization and adaptation of the pipeline across different call center domains and languages will be tested, alongside a scalability study to handle larger data volumes. Methods to incorporate multi-turn conversational context and advanced filtering and curation techniques will be developed to ensure higher quality Q\&A pairs.

In summary, while this work establishes a promising foundation for the automated generation of instructional datasets from call center audio, there is a vast scope for enhancements and future investigations that can lead to even more capable and useful LLMs in practical customer interaction applications.

\bibliographystyle{splncs04}
\bibliography{Parts/bibliography}

\end{document}